\documentclass[a4paper,12pt]{article}
\usepackage[margin=1in]{geometry}

\usepackage{lineno,hyperref}
\usepackage{authblk}

\modulolinenumbers[5]

\usepackage[square,numbers]{natbib}
\usepackage{amsmath}
\usepackage{multicol}
\usepackage{caption}
\usepackage{graphicx}
\usepackage{hyperref}
\usepackage{textcomp}
\usepackage{amsfonts}
\usepackage{floatrow}
\usepackage{makecell}
\usepackage{siunitx}
\usepackage{etoolbox}
\usepackage{pdfpages}
\usepackage{wrapfig}
\usepackage{todonotes}
\usepackage{subcaption}
\usepackage{indentfirst}

\usepackage[toc,page]{appendix}
\usepackage{chngcntr}

\usepackage{array}
\newcolumntype{P}[1]{>{\centering\arraybackslash}p{#1}}

\robustify\bfseries
\newcommand{\norm}[1]{\left\lVert#1\right\rVert}

\usepackage[ruled,vlined]{algorithm2e}

\graphicspath{{plots/}}

\begin{document}\sloppy
\title{Integrating Contrastive Learning with Dynamic Models for Reinforcement Learning from Images}

\author[1,2]{Bang You}
\author[1]{Oleg Arenz}
\author[2]{Youping Chen}
\author[1]{Jan Peters}
\affil[1]{Intelligent Autonomous Systems, TU Darmstadt}
\affil[2]{School of Mechanical Science and Engineering, Huazhong University of Science and Technology}

\maketitle





\begin{abstract}
Recent methods for reinforcement learning from images use auxiliary tasks to learn image features that are used by the agent's policy or Q-function. In particular, methods based on contrastive learning that induce linearity of the latent dynamics or invariance to data augmentation have been shown to greatly improve the sample efficiency of the reinforcement learning algorithm and the generalizability of the learned embedding. We further argue, that explicitly improving Markovianity of the learned embedding is desirable and propose a self-supervised representation learning method which integrates contrastive learning with dynamic models to synergistically combine these three objectives: (1) We maximize the InfoNCE bound on the mutual information between the state- and action-embedding and the embedding of the next state to induce a linearly predictive embedding without explicitly learning a linear transition model, (2) we further improve Markovianity of the learned embedding by explicitly learning a non-linear transition model using regression, and (3) we maximize the mutual information between the two nonlinear predictions of the next embeddings based on the current action and two independent augmentations of the current state, which naturally induces transformation invariance not only for the state embedding, but also for the nonlinear transition model. Experimental evaluation on the Deepmind control suite shows that our proposed method achieves higher sample efficiency and better generalization than state-of-art methods based on contrastive learning or reconstruction.
\end{abstract}



\section{Introduction}
Deep reinforcement learning (RL) is a promising framework for enabling robots to perform complex control tasks from high-dimensional sensory inputs in unstructured environments, including household chores~\cite{levine2016end}, manufacturing~\cite{li2019robot}, and transportation~\cite{zhu2017target}. Specifically, end-to-end reinforcement learning from images enables robots to learn new skills without relying on object-specific detection and tracking systems. However, by operating on high-dimensional observation spaces that are typically governed by complex dynamics, reinforcement learning from images often requires many environment interactions to learn successful policies, which is not possible for a wide range of real robotic tasks.

\citet{hafner2020Dream, hafner2019learning} substantially improved the sample efficiency of image-based reinforcement learning, by learning a predictive model on a learned latent embedding of the state. They alternately optimize a policy based on the learned models of the latent reward and transition distributions, and improve their models based on newly collected samples from the learned policy. However, they still need many samples to learn a model of the environment. Furthermore, the learned world model is usually inaccurate, and the resulting prediction errors lead to suboptimal policies. 

Another solution to improve sample efficiency for reinforcement learning from images is to use auxiliary tasks to learn compact state representations, which can be used by the reinforcement learning agent. One common auxiliary task is to reconstruct raw images with autoencoders and its variations~\cite{mattner2012learn, watter2015embed, yarats2019improving}. However, since such approaches learn a representation by minimizing the reconstruction error in the pixel space, they try to capture pixel-level details even when they are task-irrelevant, which can degrade sample efficiency~\cite{zhang2020learning}. 

Recently, contrastive learning objectives, which rely on mutual information estimation, have been shown to improve sample efficiency of reinforcement learning algorithms. In particular, two different types of contrastive learning have been shown to be effective for reinforcement learning from images. The first line of research~\cite{laskin2020curl} maximizes the mutual information between two independent transformations of the same image, which increases the robustness of representations similar to data augmentation. The second line of research~\cite{ anand2019unsupervised, oord2018}, maximizes mutual information between consecutive states, aiming to learn representations that have approximately linear latent dynamics. 



We further argue that improving latent Markovianity---that is, maximizing the predictability of the next embedded state based on the current embedded state and action---is important, since reinforcement learning assumes Markovian states and actions and an agent using non-Markovian representations is, thus, not able to exploit non-Markovian effects. Although mutual information based approaches that maximize predictive information~\cite{ lee2020predictive} already improve Markovianity, we argue that it is more effective to explicitly minimize the prediction errors of a learned model of the latent dynamics. 


Based on the three desired properties that we identified, we propose a self-supervised representation learning method that integrates \textbf{co}ntrastive learning with \textbf{dy}namic models, CoDy, for reinforcement learning from images. Our method consists of three auxiliary tasks on the learned embedding. Namely, (1) we minimize the prediction error between the current state- and action-embeddings and the true embedding of the next state to increase Markovianity, (2) we maximize the InfoNCE~\cite{oord2018} bound on the temporal mutual information between the current state- and action-embeddings and the true embedding of the next state to increase linearity of the latent dynamics, and (3) we maximize the multi-view mutual information between the predicted embeddings at the next time step for two independent data augmentations to improve invariance to data augmentations.

Compared to aforementioned contrastive learning methods, our proposed method offers three appealing properties. Firstly, our method effectively improves Markovianity of the latent embedding by explicitly learning a nonlinear transition models. Secondly, instead of directly maximizing the mutual information of the augmented images, we maximize the multi-view mutual information between the predicted embeddings of next states, which additionally encourages the latent transition model to be invariant to data augmentations. Thirdly, our mutual information objectives take into account the actions such that the state representation does not implicitly depend on the actions in the replay buffer.

We train our auxiliary tasks  with a standard soft actor-critic~\cite{haarnoja2018soft} reinforcement learning agent for learning continuous policies from images. The main contributions of our work are as follows.
\begin{itemize}
    \item To improve the Markovianity of state embeddings, we propose a self-supervised representation learning method that combines contrastive learning with a nonlinear prediction task. Our method learns state embeddings while inducing Markovianity, transformation invariance, and linearity of latent dynamics.
    \item We propose a novel multi-view mutual information objective that maximizes the agreement between the \emph{predicted} embeddings of the next states for different transformations of the current state, which induces transformation invariance not only for the state embeddings, but also for the latent dynamics.
    \item We evaluate our method on a set of challenging image-based benchmark tasks and show that it achieves better sample efficiency and generalization than state-of-art reconstruction-based, contrastive-learning-based and model-based methods.
\end{itemize}

The remainder of the paper is organized as follows. We present the problem statement and preliminaries in Section~\ref{Problem statement and notation}. In Section~\ref{section:Prior work} we discuss previous work related to state representation learning. We present the proposed framework and the auxiliary tasks in Section~\ref{section:method}. Section~\ref{sec:Experiment} contains details on our implementation of the proposed algorithm and the results of the experimental evaluation. In Section~\ref{sec:conclusion} we draw a conclusion and discuss limitations and future work.

\subsection{Problem Statement and Preliminaries}
\label{Problem statement and notation}
We describe the problem of learning continuous control policies from high-dimensional observations. Our algorithm is built on top of soft actor critic (SAC)~\cite{haarnoja2018soft}, which is a model-free off-policy reinforcement learning algorithm with entropy regularization. We also introduce contrastive predictive coding~\cite{oord2018} used for mutual information estimation.

\subsubsection{Problem Statement and Notation}
We formulate the problem of learning continuous control policies from images as an infinite-horizon Markov decision process (MDP). An MDP can be formulated by the tuple $\mathcal{M} =(\mathcal{S},\mathcal{A},P,r,\gamma)$, where $\mathcal{S}$ is the state space, and $\mathcal{A}$ is the action space, $P(s_{t+1}|s_t,a_t)$ is a stochastic dynamic model, $r(s,a)$ is the reward function and $\gamma$ the discount factor. The state and action space fulfill the (first-order) Markov property, that is, the distribution of the next state $s_{t+1}$ is conditionally independent of all prior states and actions, $s_{t'<t}$ and $a_{t'<t}$, given the current state $s_t$ and action $a_t$. At every time step $t$, the agent observes the current state and chooses its action based on its stochastic policy $\pi(a_t|s_t)$ and obtains a reward $r(s_t, a_t)$. Our goal is to optimize the policy to maximize the agent's expected cumulative reward. 

We specifically focus on image-based reinforcement learning, that is, the state space is provided in terms of images. A single image is usually not Markovian, since it contains little information about object velocities (which could be estimated using previous images). Following common practice~\cite{mnih2013playing} in reinforcement learning from images, we stack the $k$ most recent images together and define the state as $s_t = (o_t, o_{t-1}, o_{t-2}, \cdots, o_{t-k+1})$, where $o_t$ is the image observed at time $t$. While this problem could also be framed as a partially observable MDP (POMDP), please note that we assume that the agent observes $s_t$, which is assumed to be a Markovian state. Hence, in contrast to POMDP methods, we neither need to use previous observations $s_{t'<t}$ to better estimate an unobserved hidden state, nor do we need to learn a policy that actively chooses informative actions. Instead, we focus on representation learning, that is, we want to learn an embedding $\phi_\alpha : \mathcal{S} \rightarrow \mathbb{R}^d$, parameterized by $\alpha$, that maps the high-dimensional state $s$ to a lower-dimensional representation $z = \phi_\alpha(s)$ to increase sample efficiency. Along with the embedding $\phi_\alpha$, we want to learn a policy $\pi(a|s) = \pi(a|\phi_\alpha(s))$ that maximizes the expected cumulative rewards 
\begin{equation}
    J(\pi) = \mathbb{E}_{\pi} \left [ \sum_{t=0}^\infty \gamma ^ t r_t \bigg | a_t \sim \pi(\cdot | z_t), s_{t+1} \sim P(\cdot | s_t, a_t), s_0 \sim p_0(s_0)\right], 
\end{equation}
where $p_0(s_0)$ is the distribution of the initial state and $\gamma \in (0,1)$ is a discount factor to ensure finite returns.

\subsubsection{Maximum Entropy Reinforcement Learning}
Maximum entropy reinforcement learning optimizes a policy to maximize the sum of the expected cumulative rewards and the expected entropy of the policy~\cite{ziebart2008maximum}. Unlike standard reinforcement learning, in the maximum entropy reinforcement learning framework the agent gets an additional reward that is proportional to the entropy, $H(\pi(\cdot |s_t) = - \int_\mathcal{A} \pi(\cdot|s_t)\log(\pi(\cdot|s_t))) \, \mathrm{d}a$, of the policy at every time step to encourage stochasticity. Soft actor-critic~\cite{haarnoja2018soft} is an off-policy maximum entropy reinforcement learning algorithm that learns an entropy-regularized stochastic policy $\pi_\omega$ parameterized by $\omega$, and two $Q$ functions $Q_{\sigma_1}$ and $Q_{\sigma_2}$ with parameters $\sigma_1$ and $\sigma_2$ respectively, to find an optimal control policy. The soft $Q$ function can be learned by minimizing the Bellman error,
\begin{equation}\label{eq:Q loss}
    L_Q(\sigma_i) = \mathbb{E}_{(s,a,r,s^\prime,d) \sim D}  \left[ \left( Q_{\sigma_i}(s,a) - \mathbb{T}(r, s^\prime,d))\right)\right],
\end{equation}
where state $s$, action $a$, reward $r$, next state $s^\prime$ and a termination flag $d$ are sampled from the replay buffer $D$. The target value $\mathbb{T}(r, s^\prime, d)$ is computed from the $Q$ function
\begin{equation}
    \mathbb{T}(r, s^\prime,d) = r + \gamma(1-d) \Big( \min_{i=1,2} Q_{\sigma_{\textsubscript {targ,i}}}(s^\prime, a^\prime) - \alpha \log \pi_\omega(a^\prime | s^\prime) \Big),
\end{equation}
where $a^\prime \sim \pi_\omega(\cdot | s^\prime)$, and $Q_{\sigma_{\textsubscript {targ,i}}}$ denotes the target $Q$ function, which uses an exponential moving average of the parameters $\sigma_i$ for $i=1,2$. This particular parameter update has been shown to stabilize training.

The policy can be optimized by minimizing
\begin{equation}\label{eq:policy loss}
    L_\pi(\omega) = \mathbb{E}_{s,a \sim D,\pi}[\alpha \log \pi_\omega(a | s) - Q(s,a)],
\end{equation}
where the states $s$ and actions $a$ are sampled from the replay buffer and the stochastic policy, respectively. $Q(s,a) = \min\limits_{i=1,2} Q_{\sigma_i}$ is the minimum of both $Q$ function approximators.

\subsubsection{Contrastive Predictive Coding}
For two random variables $x_1$ and $x_2$, the mutual information $I(x_1, x_2)$ is defined as the Kullback-Leibler divergence between the joint distribution $p(x_1, x_2)$ and the product of the marginal distributions $p(x_1) p(x_2)$, 
\begin{equation*}
    I(x_1, x_2) = \int_{\mathbf{x_1},\mathbf{x_2}} p(x_1,x_2) \log \frac{p(x_1, x_2)}{p(x_1) p(x_2)} \mathrm{d}x_1\mathrm{d}x_2,
\end{equation*}
which is typically intractable to compute.

Contrastive predictive coding~\cite{oord2018} introduced the InfoNCE loss for maximizing a lower bound on the mutual information between the current embedding and a future embedding.
Specifically, given an anchor $x_{1,i}$, a positive sample $x_{2,i} \sim p(x_2 | x_{1,i})$ and K negative samples $x_{2,j} \sim p(x_2)$, the InfoNCE loss corresponds to a binary cross-entropy loss for discriminating samples from the joint distribution $p(x_{1,i}) p(x_2 | x_{1,i})$ from samples from the marginals, that is, 
\begin{equation}\label{eq:InfoNCE loss}
     L_{\textsubscript{NCE}} = - \underset{p(x_1,x_2)}{\mathbb{E}} \bigg[\log\frac{exp(f(x_{1,i}, x_{2,i}))}{\sum_{j=1}^K exp(f(x_{1,i}, x_{2,j}))} \bigg],
\end{equation}
where the learnable function $f(\cdot,\cdot)$ measures the similarity between the anchor and samples. 

Minimizing the InfoNCE loss is equivalent to maximizing a lower bound on the mutual information between $x_1$ and $x_2$,
\begin{equation}
    I(x_1,x_2) \geq \log(K) - L_{\textsubscript{NCE}}.
\end{equation}

Contrastive predictive coding uses a recurrent neural network to extract a context $m_t$ from the embeddings of previous observation, and maximizes the InfoNCE bound of the mutual information between the current context and the k-step future state embedding $z_{t+k}$. \citet{oord2018} propose a simple log-bilinear model for the classifier, that is, $$f(m_t, x_{t+k}) = \exp{\left(m_t^\top W  \phi(x_{t+k})\right)},$$ where $\phi$ is the encoder. The embedding can thus be learned by minimizing the InfoNCE loss both with respect to the parameters of the encoder $\phi$, and with respect to the matrix $W$ which parameterizes $f$. By maximizing the inner product between the context $m_t$ and a linear transformation of the future embedding $W\phi(x_{t+k})$, the InfoNCE loss with log-bilinear classifier favors linearly predictive embeddings~\cite{anand2019unsupervised}, although the mutual information objective does not impose linearity \emph{perse}.


\section{Related Prior Work}
\label{section:Prior work}
Learning expressive state representations is an active area of research in robot control and reinforcement learning. We will now briefly discuss robotic-prior based and reconstruction based approaches, but will focus on methods based on mutual information, which are most related to our work.

\subsection{Robotic Prior-based Methods}
Prior work in this area has explored using prior knowledge about the world or dynamics, called robotic priors, for learning state representations~\cite{jonschkowski2015learning,jonschkowski2017pves}.
These robotic priors can be defined as loss functions to be minimized without additional semantic labels. 
\citet{jonschkowski2015learning} use prior knowledge, such as temporal continuity or causality, to learn representations that improve the performance of reinforcement learning. \citet{lesort2019deep} propose to stabilize the learned representation by forcing two states corresponding to the same reference point to be close to each other. For example, the fixed starting position of the arm can be used as a reference point, which acts as a reference or calibration coordinate. \citet{morik2019state} combine the idea of robotic priors with LSTMs to learn task relevant state representations, which have been shown to be robust to states that are incorrectly classified as being close to each other. However, these robotic prior based approaches often rely on significant expert knowledge and hence limit the generalizability of the learned representations across tasks. 

\subsection{Reconstruction-based Approaches}
Many existing approaches use a reconstruction loss to learn a mapping from observations to state representations~\cite{mattner2012learn, finn2016deep}. Reconstruction-based methods aim to reconstruct their input under constraints on their latent representations, e.g. their dimensionality. Autoencoders can be used to encode high-dimensional inputs into a low-dimensional latent state space~\cite{mattner2012learn}. In order to improve training stability, \citet{yarats2019improving} incorporate variational autoencoders into an off-policy learning algorithm. However, there is no guarantee that the learned representation will encode useful information for the task at hand. To alleviate this problem, constraints on the latent dynamics have been proved effective in learning useful task-oriented representations~\cite{watter2015embed, van2016stable, finn2016unsupervised}. These methods encourage the encoder to capture information necessary to predict the next state. However, since they extract representations by minimizing the reconstruction error in pixel space, these reconstruction-based methods aim to capture the full diversity of the environment, even if it is irrelevant to the task.

\subsection{Mutual Information Based Approaches}
Recent literature on unsupervised representation learning focuses on extracting latent embeddings by maximizing different lower bounds on the mutual information between the representations and the inputs. Commonly used bounds include MINE~\cite{belghazi2018mine}, which estimates mutual information based on the Donsker-Varadhan representation~\cite{donsker1975asymptotic} of the Kullback-Leibler divergence, and InfoNCE~\cite{oord2018}, which uses a multi-sample version of noise-contrastive estimation~\cite{gutmann2010noise}.

\citet{hjelm2018learning} maximize mutual information between the input and its representations. They propose to maximize both, this global mutual information and local mutual information that considers small patches of the input. The local mutual information objective should encourage the encoder to capture features shared by different patches, and thus put less focus on pixel-level noise. 
\citet{bengio2017independently} considered simple reinforcement learning problems, and propose to learn independently controllable features, by learning a separate policy for every dimension of the embedding that only affects variation of that respective dimension.
\citet{anand2019unsupervised} proposed to learn state representations by maximizing the mutual information of observations across spatial and temporal axes. Their key idea is to improve temporal coherence of observations. However, they do not take into account the action for their temporal mutual information objective and hence the predictive information is specific to actions that were used in the training set.  

More recently, several works apply mutual information objectives for representation learning in a deep reinforcement learning setting.
\citet{oord2018} introduced the InfoNCE bound on the mutual information and already evaluated it in a reinforcement learning setting. They maximize the mutual information between the current context, which is computed from all previous state embeddings, and the embedding several time steps in the future.  
\citet{laskin2020curl} learn useful state representations by maximizing the mutual information between the features of independent transformations of the same observations, improving transformation invariance of the learned embedding. The importance of data augmentation for deep reinforcement learning from images has also been stressed by~\citet{yarats2021image}.  \citet{laskin2020curl} generate transformed images for data augmentation, where a random patch is cropped from a stack of temporally sequential frames sampled from the replay buffer. Moreover, they use separate online and target encoders for the anchor, and positive/negative samples, respectively, when computing the InfoNCE loss on the mutual information, rather than using same encoder for the anchor, and positive/negative samples, which was  proposed by ~\citet{oord2018}. Our method shares several aspects with their method, CURL, since we also apply the InfoNCE~\cite{oord2018} bound to maximize the mutual information between two embeddings that result from different image crops in one of our auxiliary tasks. However, we do not consider the embeddings directly, but the predicted embeddings at the next time step, additionally targeting latent dynamics invariant to the transformation of the current embedding.
\citet{lee2020predictive} apply the conditional entropy bottleneck~\cite{Fischer2020conditional} to maximize the mutual information between the current embedding and the future state and reward, while compressing away information that is not also contained in the next state or reward. Their conditional entropy bottleneck objective is conditioned on multiple future states and rewards for multi-step prediction. Instead, we propose to compress away task-irrelevant information by using data augmentation without compromising the ability to predict the \emph{embedding} of the next state.

Model-based reinforcement learning methods iteratively build a predictive model of the environment from high-dimensional images along with a policy based on that model~\cite{hafner2019learning, finn2017deep, zhang2019solar, agrawal2016learning}. For example, Dreamer~\cite{hafner2020Dream} learns models of the latent dynamics and rewards and uses them for reinforcement learning in the latent MDP. By iteratively applying the policy on the real system, new data is collected to improve the models. Some recent model-free reinforcement learning approaches learn latent dynamic models to compute intrinsic reward for solving many reinforcement learning tasks with sparse rewards~\cite{pathak2017curiosity, li2020random}. For instance,  \citet{li2020random} compute intrinsic rewards by estimating the novelty of the next state based on the prediction error of a dynamic model that is smoothly updated during training. Learning predictive dynamic models has also been shown to be promising for many robotic control tasks. For example, \citet{li2020neural} propose a neural fuzzy-based dynamic model for reliable trajectory tracking, which effectively facilitates the control performance of the wheel-legged robot. We also learn a model of the latent dynamics, but we use it only for learning better representations. However, making better use of the learned models is a natural extension for future work.

In this paper, we propose an approach for representation learning in the context of deep reinforcement learning, that is based on mutual information, without relying on robotic prior knowledge or pixel-reconstruction. Most aforementioned mutual information based methods either improve temporal coherence of observations or maximize the similarity of two independently augmented images, which cannot effectively guarantee the Markovianity of state and action embeddings. However, non-Markovian embeddings can make it harder to learn an optimal policy or Q-function, since reinforcement learning agents assume their states and actions to be Markovian. To alleviate this problem, our proposed method imposes a nonlinear dynamic model on the latent state space. Moreover, instead of directly maximizing the agreement of embedding of augmented images, we propose to maximize the mutual information between the predicted embedding at the next time step, additionally improving data augmentation invariance of the latent dynamic models. The comparative experiments presented in Section \ref{sec:Experiment} have shown that the proposed method outperforms leading reconstruction-based and contrastive-based methods on typically challenging image-based benchmark tasks.

\begin{figure*}
\centering
\includegraphics[width=\textwidth]{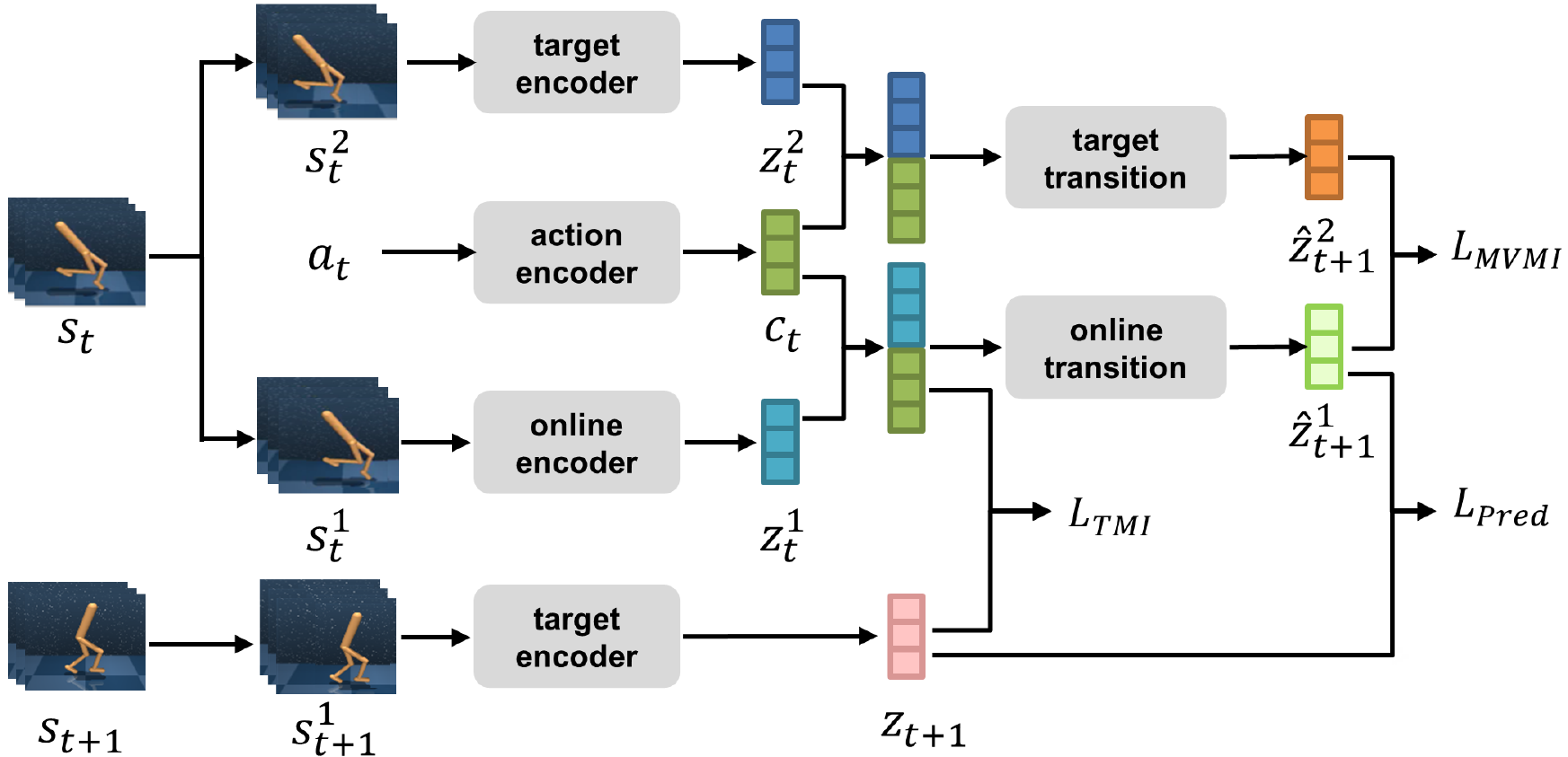}
\caption{Our framework contains three auxiliary tasks and respective loss functions. The temporal mutual information loss $L_\text{TMI}$ aims to maximize the InfoNCE bound of the mutual information between the current state-action embedding and the next state-embedding, that is, $I([c_t, z_{t}^1], z_{t+1})$. The prediction loss $L_\text{Pred}$ is given by the squared $\ell_2$ error between the predicted next embedding and the actual embedding of next state. The multi-view mutual information loss uses the InfoNCE bound to maximize the mutual information between the predicted next embeddings based on two different state augmentation $s_t^2$ and $s_t^1$, $I(\hat{z}_{t+1}^1,\hat{z}_{t+1}^2)$. 
The parameters of the target encoder and target transition function are an exponential moving average of the parameters of the online models. The online encoder is used by the Q-functions and the policy of the soft actor-critic agent, which is trained along with the auxiliary tasks.} 
\label{fig:Schematic}
\end{figure*}

\section{Integrating Contrastive Learning with Dynamic Models}
\label{section:method}

We will now present the architecture of the proposed framework and our method, which consists of three auxiliary tasks, in detail. We will also show how to train the representations together with the policy and the Q-function.

\subsection{The Mechanism and Architecture for Mutual Information Maximization}

Our mutual information maximization architecture is presented in Figure~\ref{fig:Schematic}. 
We apply stochastic data augmentation to raw states $s_t$ in order to obtain two independent views (image crops) of the current states, $s_t^1 \sim A(s_t)$ and $s_t^2 \sim A(s_t)$, where $A(\cdot)$ denotes the distribution of augmented images. In order to extract compact representations from states, we use an online encoder $\phi_\alpha : \mathcal{S} \rightarrow \mathbb{R}^d$ with parameters $\alpha$ and a target encoder  $\phi_\beta : \mathcal{S} \rightarrow \mathbb{R}^d$  with parameters $\beta$ to transform augmented observations $s_t^1$ and $s_t^2$ into representations $z_t^1$ and $z_t^2$, respectively. An action encoder $\psi_\gamma : \mathcal{A} \rightarrow \mathbb{R}^n$ with parameters $\gamma$ maps actions $a_t$ into a feature vector $c_t$. Finally, the representations of states and actions are concatenated together. An online transition model $g_\upsilon$ with parameters $\upsilon$ and a target transition model $g_\mu$ with parameters $\mu$ (e.g., neural networks) predict the representation of the next state based on a given state- and action-embedding,
\begin{align}
    \hat{z}_{t+1}^1 &= g_\upsilon(z_t^1, c_t),\\
    \hat{z}_{t+1}^2 &= g_\mu(z_t^2, c_t).
\end{align}

At the next timestep, the target encoder maps the state $s_{t+1}$ into the latent representations $z_{t+1}$. Motivated by \citet{he2020momentum}, the parameters of the target state encoder and the target transition model are computed as an exponential moving average (EMA) of the parameters of the online encoder and online transition model, respectively,

\begin{align}
    \beta &= \tau\alpha + (1-\tau)\beta, \\
    \mu &= \tau\upsilon + (1-\tau)\mu,
\end{align}
where $\tau \in [0,1)$ is the coefficient of the exponential moving average.

\subsection{A Prediction Task for Improving Latent Markovianity}
\label{sec:lpred}
The Markov assumption is critical in reinforcement learning and states that the distribution over the next state is conditionally independent of all previous states and actions, given the current state and action. When the Markov assumption is violated, the learned policy and value function in general cannot make use of the environmental information that is contained in the previous state but not in the current one. By providing the state embedding $z$ instead of the original state $s$ as input to the agent's policy and Q-function, we effectively change the state space from the agent's perspective.

Although the original state space is assumed Markovian, the learned state representation is not Markovian in general. For example, consider a simple linear system, where the state space is given by the position and velocity of a point mass, and the action corresponds to a bounded change of acceleration. While the state space is Markovian, a representation that discards the velocity is no longer Markovian, since the position at time step $t-1$ can improve our estimate of the current velocity and thus the prediction of the next position. 

However, we argue that a maximally compressed embedding of a Markovian state, that keeps sufficient information for learning an optimal policy and Q-function, should also be Markovian, which we can prove by contradiction: Assume a non-Markovian embedding of a Markovian state that is maximally compressed while keeping all information that is useful for predicting the expected cumulative reward. As the current state---due to Markovianity of the state space---contains all information for predicting the next state, and thus also for predicting the next embedded state, a non-Markovian state representation necessarily discards information that would be useful for predicting the next embedded state. Hence, either we lost information that would be useful for predicting the expected cumulative reward, or the state representation contains information that is unnecessary for predicting the expected cumulative reward, and, thus, not maximally compressed.


While latent Markovianity is a necessary condition for learning a concise and sufficient representation of the state, it is clearly not sufficient, since even a constant embedding, $\phi(s)=\text{const}$, is Markovian. However, we hypothesize that introducing an auxiliary objective to induce Markovianity improves the effectiveness of the learned embedding by improving its consistency. Strict enforcement of latent Markovianity seems challenging, but we can improve latent Markovianity in the sense that we reduce the predictive information about the next embedding that was present in previous state embeddings but not in the current one. We hypothesize that improving Markovianity by means of an auxiliary prediction tasks, improves the consistency and thereby the sample efficiency and generalizability for the learned embedding. This hypothesis is consistent with the experimental results of \citet{lee2020predictive} and \citet{anand2019unsupervised}, where auxiliary prediction tasks were shown to improve the effectiveness of the learned embedding.
We improve Markovianity by introducing the auxiliary task of predicting the next state embedding $z_{t+1}$ based on the embeddings of the current state and action, $z_t^1$ and $c_t$ and a learned non-linear transition model $g_\upsilon$. The prediction error is defined as
\begin{equation}
    L_{\textsubscript{pred}}(\alpha, \gamma, \upsilon) = \norm{z_{t+1}^1 - \hat{z}_{t+1}^1}_2^2,
\label{eq:pred}
\end{equation}
with the squared $\ell_2$ norm denoted by $\norm{\cdot}_2^2$ and the prediction $\hat{z}_{t+1}^1$. By minimizing this forward prediction error, the transition model forces the state encoder to learn predictive features.

\subsection{A Temporal Mutual Information Based Prediction Task}
\label{sec:ltmi}
Although minimizing the prediction error as discussed in Section~\ref{sec:lpred} should be more effective in improving Markovianity compared to maximizing a bound on the mutual information between embeddings of consecutive time steps, the latter approach has been shown to be very effective~\cite{lee2020predictive, anand2019unsupervised}. The better performance of methods that maximize predictive information using the InfoNCE bound with log-bilinear classifiers, may be caused by implicitly inducing linearly predictive representations~\cite{anand2019unsupervised}, or by the fact that these approaches only rely on a discriminative model~\cite{kim2018emi}. Hence, we propose to use both auxiliary tasks, the prediction task discussed in Section~\ref{sec:lpred} for inducing Markovianity more strongly, and a task based on mutual information, which induces linearly predictive representations without relying on a generative model.
Namely, we optimize the state encoder $\phi_\alpha$ and action encoder $\psi_\gamma$ to maximize the temporal mutual information, between the current state- and action-embeddings and the next state-embedding, that is,
\begin{equation}
\begin{split}
    \max_{\alpha, \gamma}I([z_t^1, c_t], z_{t+1}) = \max_{\alpha, \gamma} \mathbb{E}\bigg[\log\frac{p(z_t^1, c_t, z_{t+1})}{p(z_t^1,c_t)p(z_{t+1})} \bigg],
\end{split}
\end{equation}
where $p$ denotes the joint distribution of these variables, as well as their associated marginal distributions.

We maximize the temporal mutual information by minimizing the InfoNCE loss (Eq.~\ref{eq:InfoNCE loss}), and use a log-bilinear classifier to induce linearity.
More specifically, let $(z_t^1, c_t, z_{t+1})$ denote samples drawn from the joint distribution $p(z_t^1, c_t, z_{t+1})$ which we refer to as positive sample pairs, and $N_1$ denotes a set of negative samples sampled from the marginal distribution $p(z_{t+1})$. The InfoNCE loss for maximizing the lower bound of this temporal mutual information is given by
\begin{equation}
    \label{eq:tmi}
    L_{\textsubscript{TMI}}(\alpha, \gamma) = - \underset{p(z_t^1, c_t, z_{t+1})}{\mathbb{E}} \bigg[ \underset{N_1}{\mathbb{E}}\bigg[\log\frac{h_1(z_t^1, c_t, z_{t+1})}{\sum_{z_{t+1}^* \in N_1 \cup z_{t+1}}h_1(z_t^1, c_t, z_{t+1}^*)}\bigg] \bigg],
\end{equation}
where $h_1(z_t^1, c_t, z_{t+1})$ is a log-bilinear score function given by 
\begin{equation}
    \label{eq:logbilinearclassifier}
    h_1(z_t^1, c_t, z_{t+1}) = \exp \big(m(z_t^1, c_t)^\top W_1 z_{t+1} \big).
\end{equation}
Here, the function $m(\cdot,\cdot)$ concatenates the representations of states and actions and $W_1$ is a learned matrix. The term $m(z_t^1, c_t)W_1$ in Equation~\ref{eq:logbilinearclassifier} performs a linear transformation to predict the next state representations $z_{t+1}$, which forces the encoders of states and actions to capture linearly predictive representations.

In practice, we randomly draw a minibatch of state-action pairs and corresponding next states $(s_t, a_t, s_{t+1})$ from the replay buffer. We obtain a minibatch of positive sample pairs $(z_t^1, c_t, z_{t+1})$ by feeding $s_t$, $a_t$ and $s_{t+1}$ into their corresponding encoder, respectively. For a given positive sample pair $(z_t^1, c_t, z_{t+1})$, we construct $N_1$ by replacing $z_{t+1}$ with all features $z_{t+1}^*$ from other sample pairs $(z_t^{1^*}, c_t^*, z_{t+1}^*)$ in the same minibatch.

\subsection{A Mutual Information Based Task for Improving Transformation Invariance}
The temporal mutual information objective presented in Section~\ref{sec:ltmi} can encourage the encoder of states and actions to capture task-irrelevant information which may help in discriminating the positive and negative samples in the InfoNCE loss (Eq.~\ref{eq:tmi}). \citet{lee2020predictive} propose to learn a more compressed representation by applying the conditional entropy bottleneck, which additionally punishes the conditional mutual information between the past states and the embedding, conditioned on the future. However, in general, information that is no longer available at the next time step might still be relevant for the agent at the current step. Instead, we use data augmentation to add an inductive bias on which information is not relevant for the agent, which has been shown crucial by \citet{yarats2021image}. 

For example, some lower-level factors presented in the observed states and actions, including noise or external perturbations, are typically not useful for the agent. For our experiments, we consider a multi-view setting, inspired by~\citet{laskin2020curl}, where we assume each view of the same scene shares the same task-relevant information while the information not shared by them is task-irrelevant. Intuitively, this assumption represents an inductive bias that the way we see the same scene should not affect the internal state of the environment. Instead, the representations should extract task-relevant information that is shared by different views for predicting the next embedding. Hence, we propose to maximize the multi-view mutual information $I(\hat{z}_{t+1}^1, \hat{z}_{t+1}^2)$ between the predicted representation of next states $\hat{z}_{t+1}^1$ and $\hat{z}_{t+1}^2$, based on independent views $s_t^1$ and $s_t^2$ of the same scene with respect to the state embedding $\phi_\alpha$, the action embedding $\psi_\gamma$ and the nonlinear transition model $g_\upsilon$,
\begin{equation} \label{eq:MVMI}
\begin{split}
     \max_{\alpha, \gamma, \upsilon}I(\hat{z}_{t+1}^1, \hat{z}_{t+1}^2)  = \max_{\alpha, \gamma, \upsilon} \mathbb{E} \bigg[\log\frac{p(\hat{z}_{t+1}^1,\hat{z}_{t+1}^2)}{p(\hat{z}_{t+1}^1)p(\hat{z}_{t+1}^2)} \bigg].
\end{split}
\end{equation}

Since mutual information measures the amount of information shared between random variables, this objective forces the encoder of states to capture transformation invariant information about higher-level factors (e.g. presence of certain objects) from multiple views of the same scene. Notably, instead of modelling the mutual information between the representations of two augmented states $I(z_t^1, z_t^2)$ \cite{laskin2020curl}, the mutual information objective that we consider imposes transformation invariance not only on the state embedding, but also on the transition model.

We employ InfoNCE to maximize a lower bound of the above multi-view mutual information objective. By $(\hat{z}_{t+1}^1, \hat{z}_{t+1}^2)$ we denote samples drawn from the joint distribution $p(\hat{z}_{t+1}^1, \hat{z}_{t+1}^2)$ which we refer to positive sample pairs, and by $N_2$ we denote a set of negative samples sampled from the marginal distribution $p(\hat{z}_{t+1}^2)$. 
The multi-view mutual information objective is thus given by
\begin{equation}
    L_{\textsubscript{MVMI}}(\alpha, \gamma, \upsilon) = - \underset{p(\hat{z}_{t+1}^1, \hat{z}_{t+1}^2)}{\mathbb{E}}\bigg[ \underset{N_2}{\mathbb{E}}\bigg[\log\frac{h_2(\hat{z}_{t+1}^1, \hat{z}_{t+1}^2)}{\sum_{\hat{z}_{t+1}^{2^*} \in N_2 \cup \hat{z}_{t+1}^2}h_2(\hat{z}_{t+1}^1, \hat{z}_{t+1}^{2^*})}\bigg]\bigg],
\label{eq:mv}
\end{equation}
with score function $h_2(\cdot,\cdot)$ which maps feature pairs onto scalar-valued scores. We employ a log-bilinear model,
\begin{equation}
    h_2(\hat{z}_{t+1}^1, \hat{z}_{t+1}^2) = \exp \big(\hat{z}_{t+1}^1 W_2 \hat{z}_{t+1}^2 \big),
\end{equation}
with weight transformation matrix $W_2$. Minimizing $L_{\textsubscript{MVMI}}$ with respect to $\phi_\alpha$, $\psi_\gamma$, $g_\upsilon$ and $W_2$ maximizes the mutual information between the predicted representations of next states. In practice, we randomly sample a minibatch of state-action pairs and corresponding next states $(s_t, a_t, s_{t+1})$ from the replay buffer. We obtain positive sample pairs $(\hat{z}_{t+1}^1, \hat{z}_{t+1}^2)$ by feeding the above minibatch into our mutual information framework. For a given positive sample pair $(\hat{z}_{t+1}^1, \hat{z}_{t+1}^2)$, we construct $N_2$ by replacing $\hat{z}_{t+1}^2$ with all features $\hat{z}_{t+1}^{2^*}$ from other sample pairs $(\hat{z}_{t+1}^{1^*}, \hat{z}_{t+1}^{2^*})$ in the same minibatch.\par

Finally, the total loss function $L_{\textsubscript{CoDy}}$ for the auxiliary tasks consists of the prediction loss $L_{\textsubscript{pred}}$, the temporal mutual information loss  $L_{\textsubscript{TMI}}$ and multi-view mutual information loss $L_{\textsubscript{MVMI}}$ weighted by hyperparameters $\lambda$ and $\eta$
\begin{equation}\label{eq:total}
    L_{\textsubscript{CoDy}}(\alpha, \gamma, \upsilon) = L_{\textsubscript{MVMI}}(\alpha, \gamma, \upsilon) + \lambda L_{\textsubscript{TMI}}(\alpha, \gamma) + \eta L_{\textsubscript{pred}}(\alpha, \gamma, \upsilon).
\end{equation}
The online encoder $\phi_\alpha$, the action encoder $\psi_\gamma$ and online transition model $g_\upsilon$ is optimized simultaneously by minimizing this total loss function.

\begin{algorithm*}[ht]
\SetAlgoLined
 
 \textbf{Require:} parameters $\alpha$, $\upsilon$, $\beta$, $\mu$, $\gamma$, $\omega$, $\sigma$, $\hat \sigma$, batch size $B$, replay buffer $D$, learning rates $\rho_e$, $\rho_a$ and $\rho_c$\\
 initialize replay buffer $D$\\
 \For{each training step}{
  collect experience $(s_t, a_t, r_t, s_{t+1})$ and add it to the replay buffer $D$\\
  \For{each gradient step}{
   \textbf{Sample a minibatch of tuple: $\{s_t, a_t, r_t, s_{t+1}\}_1^B \sim D$}\\
   \textbf{Update soft Q-function:}\\
   $\{\sigma_i, \alpha \} \leftarrow \{\sigma_i, \alpha \} - \rho_c \hat{\nabla}_{\sigma_i} L_Q(\sigma_i)$ for $i \in \{1,2\}$\\
   \textbf{Update policy:}\\
   $\omega \leftarrow \omega - \rho_a \hat{\nabla}_{\omega} L_\pi(\omega)$\\
   \textbf{Update online encoder, online transition model and action encoder:}\\
   $\{\alpha, \upsilon, \gamma\} \leftarrow \{\alpha, \upsilon, \gamma\} - \rho_e \hat{\nabla}_{\{\alpha, \upsilon, \gamma\}}  L_\textsubscript{CoDy}(\alpha, \upsilon, \gamma)$\\
   \textbf{Update target Q-function:}\\
   $\hat\sigma_i \leftarrow \tau\sigma_i + (1-\tau_Q)\hat\sigma_i$ for $i \in \{1,2\}$\\
   \textbf{Update target encoder and transition model:}\\
   $\beta = \tau\alpha + (1-\tau_e)\beta$ \\
   $\mu = \tau\upsilon + (1-\tau_e)\mu$
  }
 }
 \caption{Training Algorithm for CoDy}
 \label{alg:CODY}
\end{algorithm*}
\subsection{Joint Policy and Auxiliary Task Optimization}

We train our auxiliary tasks jointly with SAC, a model-free off-policy reinforcement learning agent, by adding Eq.~\ref{eq:total} as an auxiliary objective during policy training. The policy takes the representations computed with the online encoder to choose what action to take, and to approximate Q-values. Since the reward function can provide some task-relevant information, we allow the gradient of the Q-function to back-propagate through the online encoder in order to further capture task-relevant representations of observations. We do not backpropagate the actor loss through the embedding because this degrades performance by implicitly changing the Q-function during the actor update, as noted by \citet{yarats2019improving}.

The training procedure is presented in Algorithm~\ref{alg:CODY} in detail. $\alpha$ and $\upsilon$ are the parameters of the online encoder $\phi_\alpha$ and transition model $g_\upsilon$, respectively. $\beta$ and $\mu$ are the parameters of the target encoder $\phi_\beta$ and transition model $g_\mu$, while $\gamma$ are the parameters of the action encoder $\psi_\gamma$. The parameters of policy $\pi$, the Q-function and target Q-function are denoted as $\omega$, $\sigma$ and $\hat \sigma$, respectively. $\rho_a$ and $\rho_c$ are the learning rates for the policy and Q-function. $\rho_e$ is the learning rate for the online encoder, online transition model and action encoder. The experience is stored in a replay buffer $D$, which is initialized with tuples $\{s_t, a_t, r_t, s_{t+1}\}$ by using a random policy. The algorithm proceeds by alternating between collecting new experience from the environment, and updating the parameters of the soft Q-function, policy and auxiliary prediction model. The parameters of the policy network and Q-function network are optimized by minimizing the SAC policy loss (Eq.\ref{eq:policy loss}) and actor loss (Eq.\ref{eq:Q loss}), respectively. The parameters of the online encoder, online transition model and action encoder are optimized by minimizing Eq.~\ref{eq:total} jointly. The parameters of the target Q-function network are given by an exponential moving average of the parameters of the online Q-function. 

\section{Experimental Evaluation}
\label{sec:Experiment}

We evaluate the data efficiency and performance of our method and compare it against state-of-the-art methods---both a model-based reinforcement learning method and model-free reinforcement learning methods with auxiliary tasks---on various benchmark tasks. We test the generalization of our method to unseen tasks and compare it with other methods. We will now describe the experimental setup and present the results. We will also show the results of an ablation study to show the effects of the different auxiliary tasks and a visualization of the learned representations. Our code is open-sourced and available at \mbox{\url{https://github.com/BangYou01/Pytorch-CoDy}}. 

\begin{figure*}
\centering
\includegraphics[width=\textwidth]{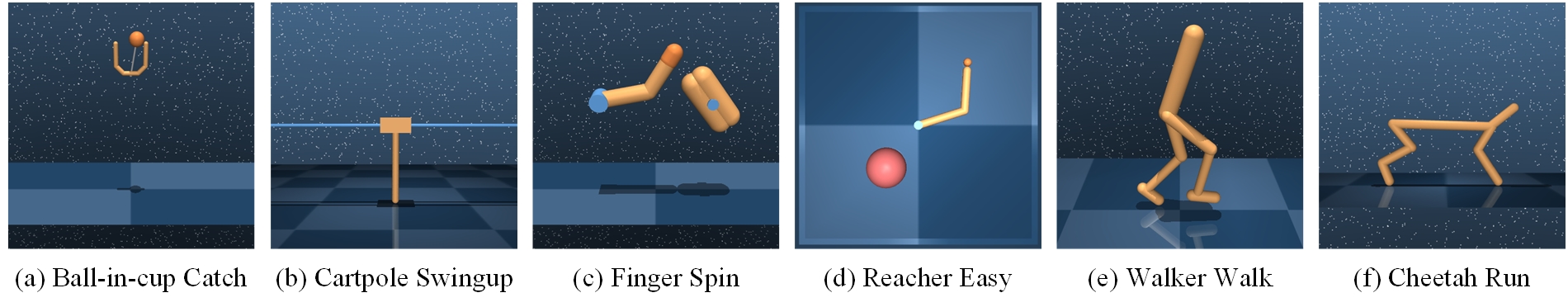}
\caption{Continuous control tasks from the Deepmind control suite used in our experiments. (a) The Ball-in-cup Catch task only provides the agent with a sparse reward when the ball is caught. (b) The Cartpole Swingup task attains images from a fixed camera and, hence, the cart can move out of sight. (c) The Finger Spin task requires contacts between the finger and the object. (d) The Reacher Easy task has a sparse reward that is only given when the target location is reached. (e) The Walker Walk task has complex dynamics. (f) The Cheetah Run task has both high action dimensions and contacts with the ground.}
\label{fig:dmc_task}
\end{figure*}
\subsection{Experimental Setup}

We implement the proposed algorithm on the commonly used PlaNet~\cite{hafner2019learning} test bed, which consist of a range of challenging imaged-based continuous control tasks (see Figure \ref{fig:dmc_task}) from the Deepmind control suite~\cite{tassa2018deepmind}. Specifically, six tasks are considered: Ball-in-cup Catch, Cartpole Swingup, Reacher Easy, Finger Spin, Walker Walk and Cheetah Run. Every task offers a unique set of challenges, including sparse rewards, complex dynamics as well as contacts. We refer to~\citet{tassa2018deepmind} for more detailed descriptions.

We parameterize the online encoder of states, the action encoder and the transition model using feed forward neural networks. The online encoder of observations consists of four convolution layers following a single fully-connected layer. We use a kernel of size $3\times 3$ with 32 channels and set stride to 1 for all convolutional layers. We employ ReLU activations after each convolutional layer. The output of the convolutional neural network is fed into a single fully-connected layer with 50-dimensional output. The action encoder consists of two fully-connected layers  with ReLU activation functions. The hidden dimension is set to 512 and the output dimension is set to 16 for the action encoder. The transition model uses three fully-connected layers with ReLU activation functions. Its hidden dimension is set to 1024 and the output dimension is set to 50.  The target encoder and transition model share the same network architecture with the online encoder and transition model, respectively. We stack 3 consecutive frames as an observation input, where each frame is an RGB rendering image with size $3 \times 84 \times 84$. We follow \citet{lee2020predictive} by augmenting states by randomly shifting the image by $[-4, 4]$. We set $\lambda = 100$ and $\eta=1000$ for all tasks.

We use the publicly released implementation of SAC by~\citet{yarats2019improving}. The Q-function consists of three fully-connected layers with ReLU activation functions. The policy network is also parameterized as a 3-layer fully-connected network that outputs the mean and covariance for a Gaussian policy. The hidden dimension is set to 1024 for both the Q-function and policy network. Table~\ref{table:SAC hyper} presents the remaining hyperparameters in detail. Following common practice~\cite{hjelm2018learning, yarats2019improving, laskin2020curl}, we treat action repeat as a hyperparameter to the agent. The number of repeated actions and the batch size for each task are listed in Table~\ref{table:per-task hyper}. 

\begin{table}
    \RawFloats
    \begin{minipage}{.5\linewidth}
      \caption{Shared hyperparameters used for the comparative experiments}
      \label{table:SAC hyper}
      \centering
        \begin{tabular}{c|P{10mm}}
            \Xhline{2\arrayrulewidth}
            Parameter & Value\\
            \hline
            Replay buffer capacity & 100000\\
            Initial steps & 1000\\
            Optimizer & Adam\\
            Q-function EMA $\tau_Q$ & 0.01\\
            Critic target update freq & 2\\
            Learning rate $(\rho_a, \rho_c)$ & $10^{-3}$ \\
            Learning rate $(\rho_e)$ & $10^{-5}$ \\
            Encoder EMA $\tau_e$ & 0.05 \\
            Discount & 0.99\\
            Initial temperature & 0.1\\
            \Xhline{2\arrayrulewidth}
        \end{tabular}
    \end{minipage}%
    \hfil
    \begin{minipage}{.5\linewidth}
      \centering
        \caption{Per-task hyperparameters}
        \label{table:per-task hyper}
        \begin{tabular}{c|P{10mm}|P{10mm}}
            \Xhline{2\arrayrulewidth}
            Task & Action \newline Repeat & Batch \newline size\\
            \hline
            Ball-in-cup Catch & 4 & 256 \\
            Finger Spin & 2 & 256\\
            Reacher Easy & 4 & 256\\
            Cartpole Swingup & 8 & 256\\
            Walker Walk & 2 & 256\\
            Cheetah Run & 4 & 512\\
            \Xhline{2\arrayrulewidth}
        \end{tabular}
    \end{minipage} 
\end{table}

We compare our algorithm with the following leading baselines for continuous control from images: CURL~\cite{laskin2020curl}, which combines a model-free RL agent with a contrastive learning objective that captures mutual information between two augmented images, PISAC~\cite{lee2020predictive}, which maximizes the predictive information between the past and the future to learn latent representations which are used by a model-free RL agent, SAC+AE~\cite{yarats2019improving}, which uses a regularized autoencoder for learning a mapping from high-dimensional states to compact embeddings, Dreamer~\cite{hafner2020Dream}, a model-based reinforcement learning method which learns a predictive dynamic model for planning, and Pixel SAC which uses a vanilla SAC operating purely from pixels. 

We evaluate the performance of every agent after every 10K environment steps by computing an average return over 10 episodes. For each method, the SAC agent performs one gradient update per environment step to ensure a fair comparison. For a more reliable comparison, we run each algorithm with five different random seeds for each task. All figures show the average reward and 95\% confidence interval unless specified otherwise.

\begin{figure*}[t]
\centering
\includegraphics[width=\textwidth]{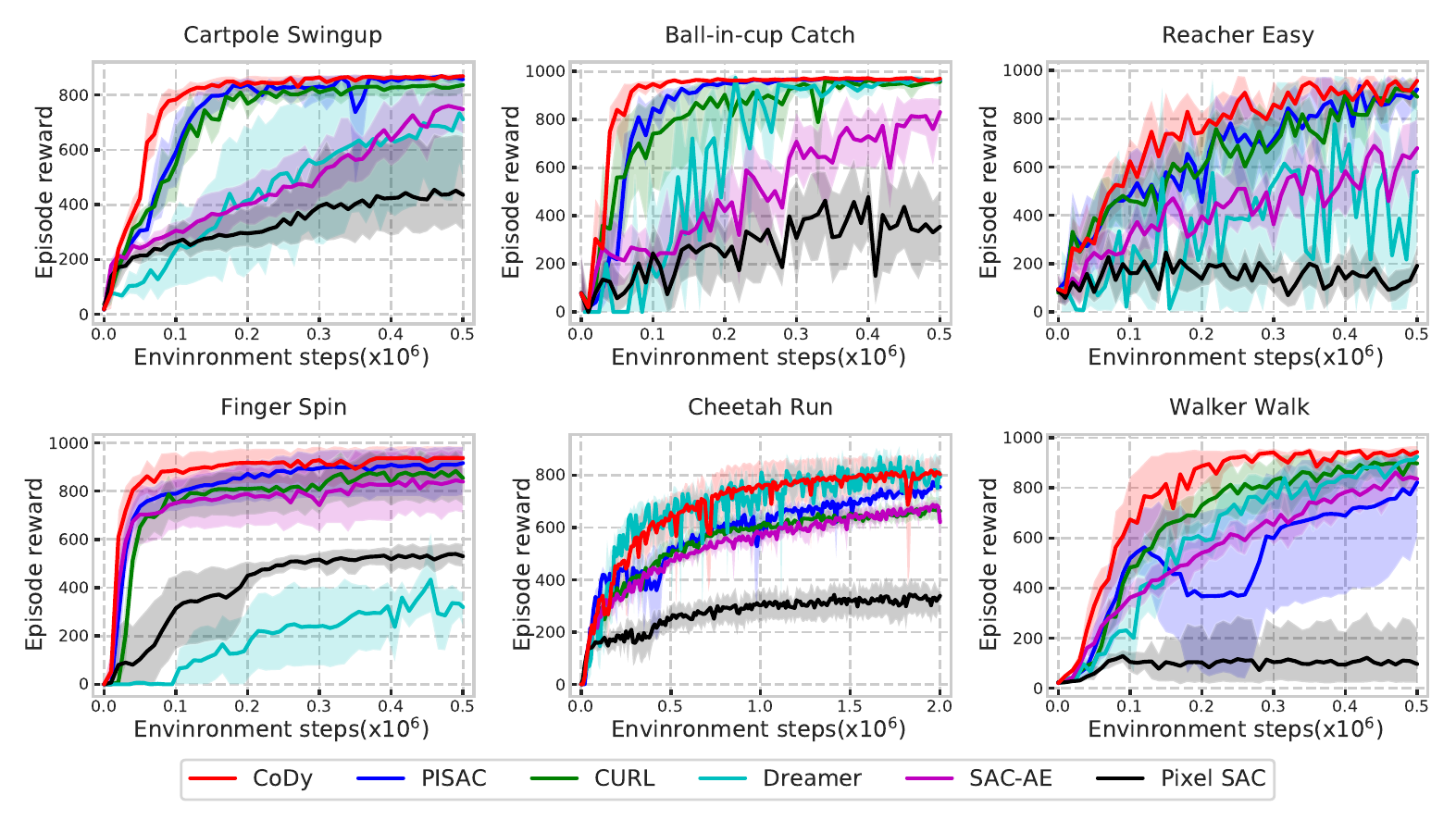}
\caption{We compared the performance of CoDy with existing methods on six tasks from the Deepmind control suite. On all tasks, CoDy performs best in terms of sample efficiency.}
\label{fig:sample_efficiency}
\end{figure*}

\subsection{Sample Efficiency}

\begin{table*}[b!]
\begin{center}
\sisetup{%
            table-align-uncertainty=true,
            separate-uncertainty=true,
            detect-weight=true,
            detect-inline-weight=math
        }
\resizebox{\textwidth}{!}{\begin{tabular}{c | c c c c c c | c}
    \Xhline{2\arrayrulewidth}
    500K step scores & CoDy(Ours) & PISAC & CURL & SAC-AE & Dreamer & Pixel SAC & State SAC\\
    \hline
    Finger Spin & \bfseries 937$\pm$ \bfseries 41$^\ast$ & \bfseries 916 $\pm$ \bfseries 58 & \bfseries 854$\pm$ \bfseries 48 & \bfseries 839$\pm$ \bfseries 68 & 320$\pm$35 & 530$\pm$24 & 927 $\pm$ 43\\
    Cartpole Swingup & \bfseries 869$\pm$ \bfseries 4$^\ast$ & \bfseries 857 $\pm$ \bfseries 12 & 837$\pm$ 15 & 748$\pm$47 & 711$\pm$ 94 & 436$\pm$94 & 870 $\pm$ 7\\
    Reacher Easy & \bfseries 957$\pm$ \bfseries 16$^\ast$ & \bfseries 922 $\pm$ \bfseries 32 & 891$\pm$ 30 & 678$\pm$61 & 581$\pm$ 160 & 191$\pm$40 & 975 $\pm$ 5\\
    Cheetah Run & \bfseries 656$\pm$ \bfseries 43$^\ast$ &  510 $\pm$  27 & 492$\pm$22 & 476$\pm$ 22 & \bfseries 571$\pm$ \bfseries 109 & 250$\pm$26 & 772 $\pm$ 60\\
    Walker Walk & \bfseries 943$\pm$ \bfseries 17$^\ast$ & 822 $\pm$ 98 & 897$\pm$ 26 & 836$\pm$24 & \bfseries 924$\pm$ \bfseries 35	& 97$\pm$62 & 964 $\pm$ 8\\
    Ball-in-cup Catch & \bfseries 970$\pm$ \bfseries 4$^\ast$ & 961 $\pm$ 3 & 957$\pm$ 6 & 831$\pm$25 & \bfseries 966$\pm$ \bfseries 8 & 355$\pm$77 & 979 $\pm$ 6\\
    \hline
    100K step scores & & & & & & & \\
    \hline
    Finger Spin & \bfseries 887 $\pm$ \bfseries 39$^\ast$ &  789 $\pm$ 34 & 750 $\pm$  37 & 751$\pm$  57 & 33 $\pm$ 19 & 315 $\pm$ 78 & 672 $\pm$ 76\\
    Cartpole Swingup & \bfseries 784 $\pm$ \bfseries 18$^\ast$ & 591 $\pm$  70 & 547 $\pm$ 73 & 305$\pm$ 17 & 235$\pm$ 73 & 263 $\pm$ 27 & 812 $\pm$ 45\\
    Reacher Easy & \bfseries 624 $\pm$ \bfseries 42$^\ast$ & 482 $\pm$  91 & 460 $\pm$ 65 & 321 $\pm$ 26 & 148 $\pm$ 53 & 160 $\pm$ 48 &  919 $\pm$ 123\\
    Cheetah Run & \bfseries 323 $\pm$ \bfseries 29$^\ast$ & \bfseries 310 $\pm$ \bfseries 28 &  266 $\pm$  27 & 264$\pm$  12 & 159$\pm$ 60 & 160 $\pm$ 13 & 228 $\pm$ 95\\
    Walker Walk & \bfseries 673 $\pm$ \bfseries 94$^\ast$ & \bfseries 518 $\pm$ \bfseries 70 &  482 $\pm$  28 & 362 $\pm$ 22 & 216 $\pm$ 56 & 105 $\pm$ 20 & 604 $\pm$ 317 \\
    Ball-in-cup Catch & \bfseries 948 $\pm$ \bfseries 6$^\ast$ & 847 $\pm$ 21 & 741 $\pm$  102 & 222 $\pm$ 21 & 172 $\pm$ 96 & 244 $\pm$ 55 & 957 $\pm$ 26\\
    \Xhline{2\arrayrulewidth}
\end{tabular}}
\end{center}
\caption{Scores achieved by our method (mean and standard error for 5 seeds) and baselines at 100k and 500k environment steps. $^\ast$ indicates the best average return among these methods. The bold font indicates that the upper-bound on the reward (based on the given standard error intervals) for the given method, is larger or equal than the respective lower bound for every other image-based method.}
\label{table:sample efficiency}
\end{table*}

Figure \ref{fig:sample_efficiency} compares our algorithm with PISAC~\cite{lee2020predictive}, CURL~\cite{laskin2020curl}, SAC-AE~\cite{yarats2019improving}, Dreamer~\cite{hafner2020Dream} and Pixel SAC~\cite{haarnoja2018soft}. We use the version of PISAC that uses the same implementation of the SAC algorithm~\cite{yarats2019improving} to ensure a fair comparison to other model-free approaches. The evaluation data of Dreamer was provided to us by the author. The proposed algorithm achieves state-of-the-art performance on all the tasks against all the leading baselines, both model-based and model-free. 

Following CURL~\cite{laskin2020curl}, in Table~\ref{table:sample efficiency} we also compare performance at a fixed number of environment interactions (100k and 500K). Dreamer's results provided by the author didn't show the performance of the agent at exactly 100k and 500K environment interactions, and hence we interpolated between nearby values. We compare our algorithm with above baselines and an upper bound performance achieved by SAC~\cite{haarnoja2018soft} that operates directly from internal states instead of images. Averaged over 5 seeds, our algorithm achieves better sample-efficiency at 100K environment interactions and asymptotic performance at 500K environment interactions against existing baselines on all tasks. Furthermore, our algorithm matches the upper bound performance of SAC that is trained directly from internal states on several tasks at 500K environment interactions.

\begin{figure*}[t]
\centering
\includegraphics[width=\textwidth]{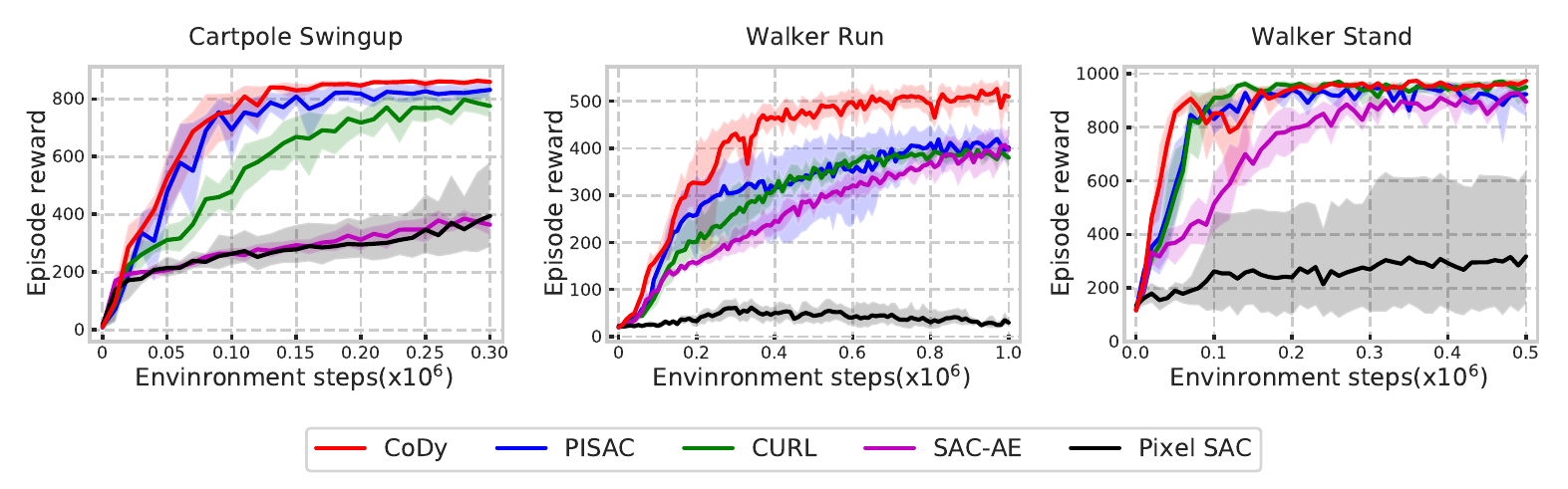}
\caption{Generalization comparisons on the Deepmind control suite. Generalization of an encoder trained on the Cartpole Balance task and evaluated on the unseen Cartpole Swingup task (left). Generalization of an encoder trained on the Walker Walk task and evaluated on the unseen Walker Run (centre) and Walker Stand (right) task, respectively.}
\label{fig:Generalization}
\end{figure*}

\begin{figure*}[t]
\centering
\includegraphics[width=\textwidth]{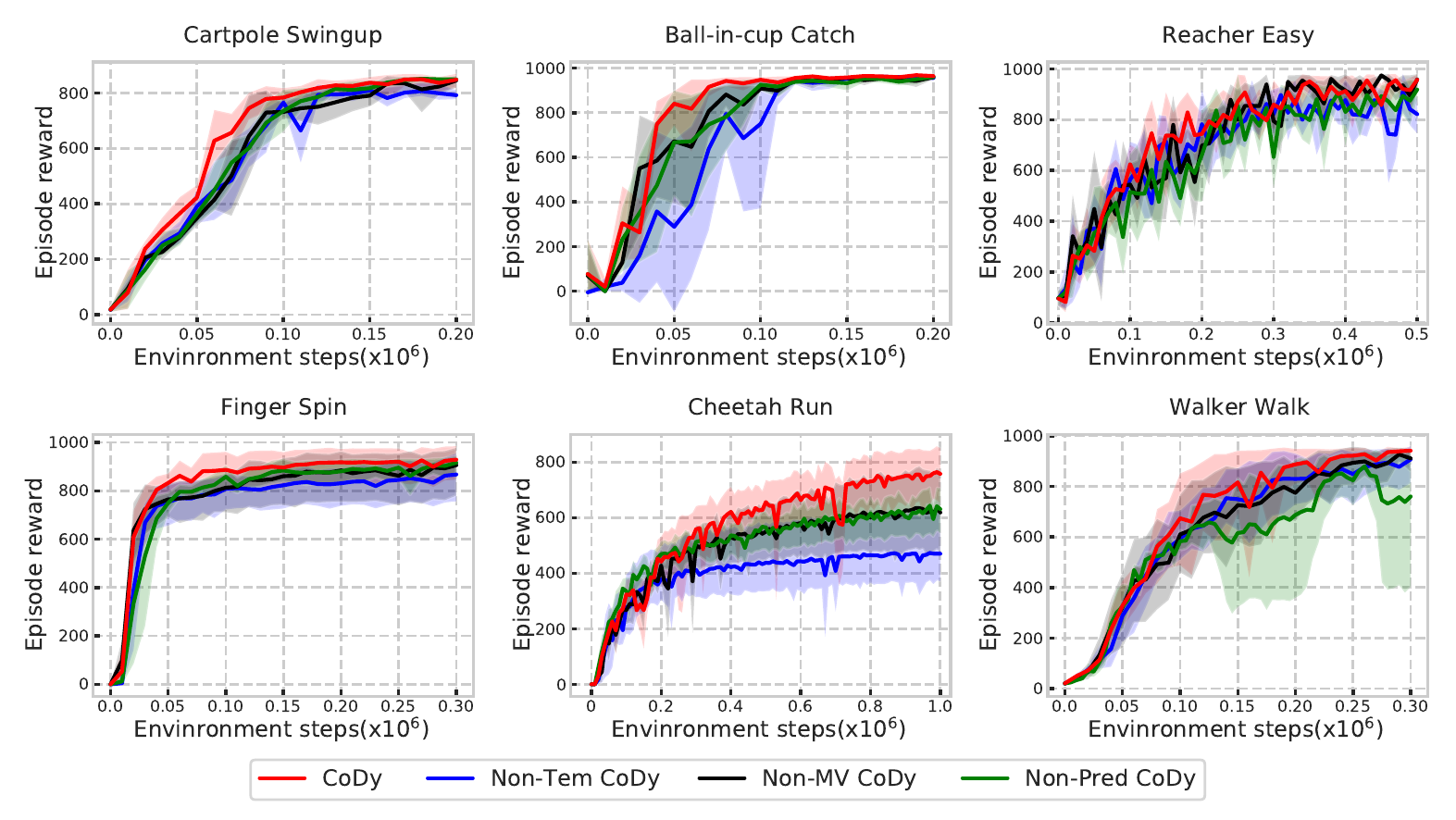}
\caption{Performance on Deepmind control suite under consideration for ablated variants of our method.}
\label{fig:ablation}
\end{figure*}

\subsection{Generalization to Unseen Tasks}

We test the generalization of our method by transferring the learned representations without additional fine-tuning to unseen tasks that share the same environment dynamics. Specifically, we learn representations and a SAC agent on a source task, fix the representations and then train a new SAC agent for a target task using the fixed representations. We use Cartpole Balance and Cartpole Swingup as the source task and the target task, respectively, which share the same environment dynamics but differ in their initial pole positions. We also use Walker Walk as the source task and Walker Stand and Walker Run as the target task, which all have different reward functions. Figure \ref{fig:Generalization} compares the generalization of our method against PISAC, CURL, SAC-AE and a Pixel SAC that was trained from scratch. Our algorithm achieves not only better generalization than baselines, but also much better performance than vanilla Pixel SAC trained from scratch.
We attribute the improved generalization compared to other contrastive learning based methods to our additional focus on improving latent Markovianity. The source and target tasks differ only in their reward function or the inital position, and share the same state-action dynamics. A Markovian representation is consistent with the state-action dynamics, in the sense that it does not discard information that would be useful for predicting aspects of the state that it chose to encode. We conjecture that this consistency improves the generalization of CoDy to unseen tasks.   


\subsection{Ablation Studies}
We perform ablation studies to disentangle the individual contributions of the multi-view objective and the temporal mutual information objective, as well as the role of the prediction loss. We investigate three ablations of our CoDy model: Non-Tem CoDy, which only optimizes multi-view mutual information objective and the prediction error; Non-Pred CoDy, which only optimizes the multi-view and temporal mutual information objective, and Non-MV CoDy, which only optimizes the temporal mutual information objective and the prediction error. We present the performance of these ablations in Figure~\ref{fig:ablation}. CoDy achieves better or at least comparable performance and sample efficiency to all its own ablations across all tasks. This indicates that all three auxiliary tasks in CoDy play an important role in improving the performance on the benchmark tasks. Notably, CoDy outperforms Non-Pred CoDy across all tasks, which indicates that improving the Markovianity of embedding by minimizing the prediction error of dynamic models effectively achieves better sample efficiency. By comparing the performance of CoDy and Non-MV CoDy, we observe that our multi-view mutual information objective based on predicted embeddings helps to achieve better performance.


\begin{figure*}[t]
\centering
\includegraphics[width=\textwidth]{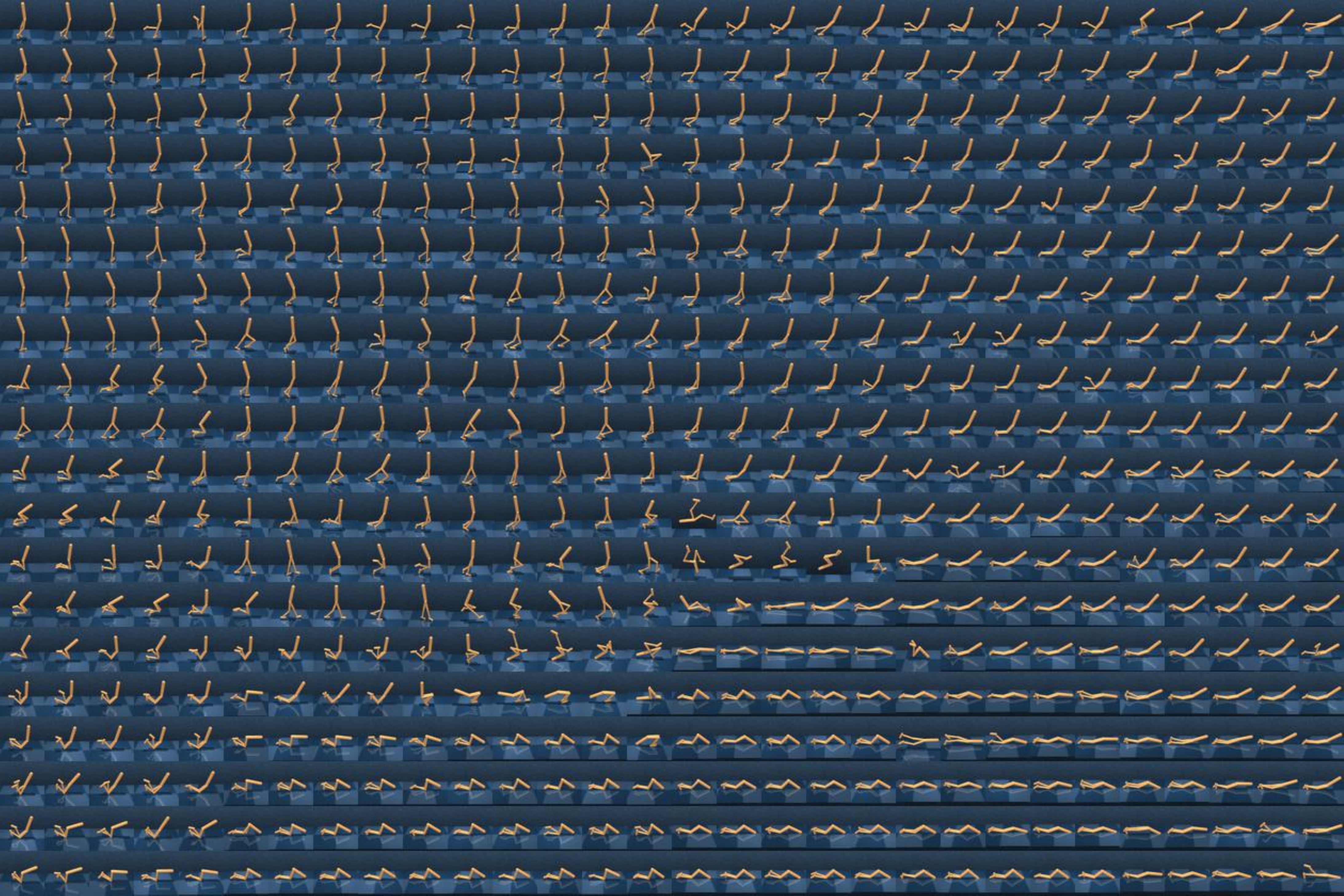}
\caption{We visualize the representations learned with CoDy after training has completed on the Walker Walk task using t-SNE Visualization on a $30 \times 20$ grid.}
\label{fig:tsne}
\end{figure*}

\subsection{Representation Visualization}
\label{sec:vis}
We visualize the representations learned by CoDy using t-SNE~\cite{van2008visualizing} to inspect the learned embedding. With t-SNE visualization, there tend to be many overlapping points in the 2D space, which makes it difficult to view the overlapped representation examples. Therefore, we quantize t-SNE points into a 2D grid with a $30\times20$ interface by RasterFairy~\cite{Klingemann2015Raster}. Figure \ref{fig:tsne} visualizes the representations learned with CoDy after training has completed on the Walker Walk task. Observations with similar robot configurations appear close to each other, which indicates that the latent space learned with CoDy meaningfully organizes the variation in robot configurations. Similar visualization for the remaining 5 tasks from the Deepmind control suite are shown in Appendix \ref{Additive Representation Visualization}.

\section{Discussion and Conclusion}
\label{sec:conclusion}
We presented CoDy, a self-supervised representation learning method which integrates contrastive learning with dynamic models for improving the sample efficiency of model-free reinforcement learning agents. Our method aims to learn state representations that are Markovian, linearly predictive and transformation invariant by minimizing three respective auxiliary losses during reinforcement learning from images. We compared our method with state-of-the-art approaches on a set of challenging image-based benchmark tasks. The results showed that our method can achieve better sample efficiency and performance compared to all the leading baselines on the majority of tasks, including  reconstruction-based, contrastive-based as well as model-based methods. Furthermore, we found representations learned with our method can achieve better generalization to unseen tasks than other model-free methods.

Although our method performed better than previous methods in our experiments, it is also a bit more complex by using three auxiliary tasks. We propose to learn a transition model in latent space and showed that it can be helpful for learning compact state embeddings, however, we do not make direct use of this model during reinforcement learning, which seems wasteful. Making use of the learned transition model for planning is a natural extension for future research. Moreover, we plan to extend our method to multimodal data, which incorporates other modalities, such as tactile and sound, into our representation learning model.

    

\section*{Declaration of Competing Interest}
The authors declare that they have no known competing financial interests or personal relationships that could have appeared to influence the work reported in this paper.

\section*{Acknowledgement}
This work was supported by the 3rd Wave of AI project from Hessian Ministry of Higher Education, Research, Science and the Arts. The financial support provided by China Scholarship Council Scholarship program (No.202006160111) is acknowledged.

\section*{Appendix}
\appendix
\counterwithin{table}{section} 
\counterwithin{figure}{section}
\addcontentsline{toc}{chapter}{Appendices}
\setcounter{section}{0}
\renewcommand{\thesection}{\Alph{section}}

\section{Additional Visualizations of the Learned Representation}
\label{Additive Representation Visualization}
\begin{figure*}
\centering
\includegraphics[width=\textwidth]{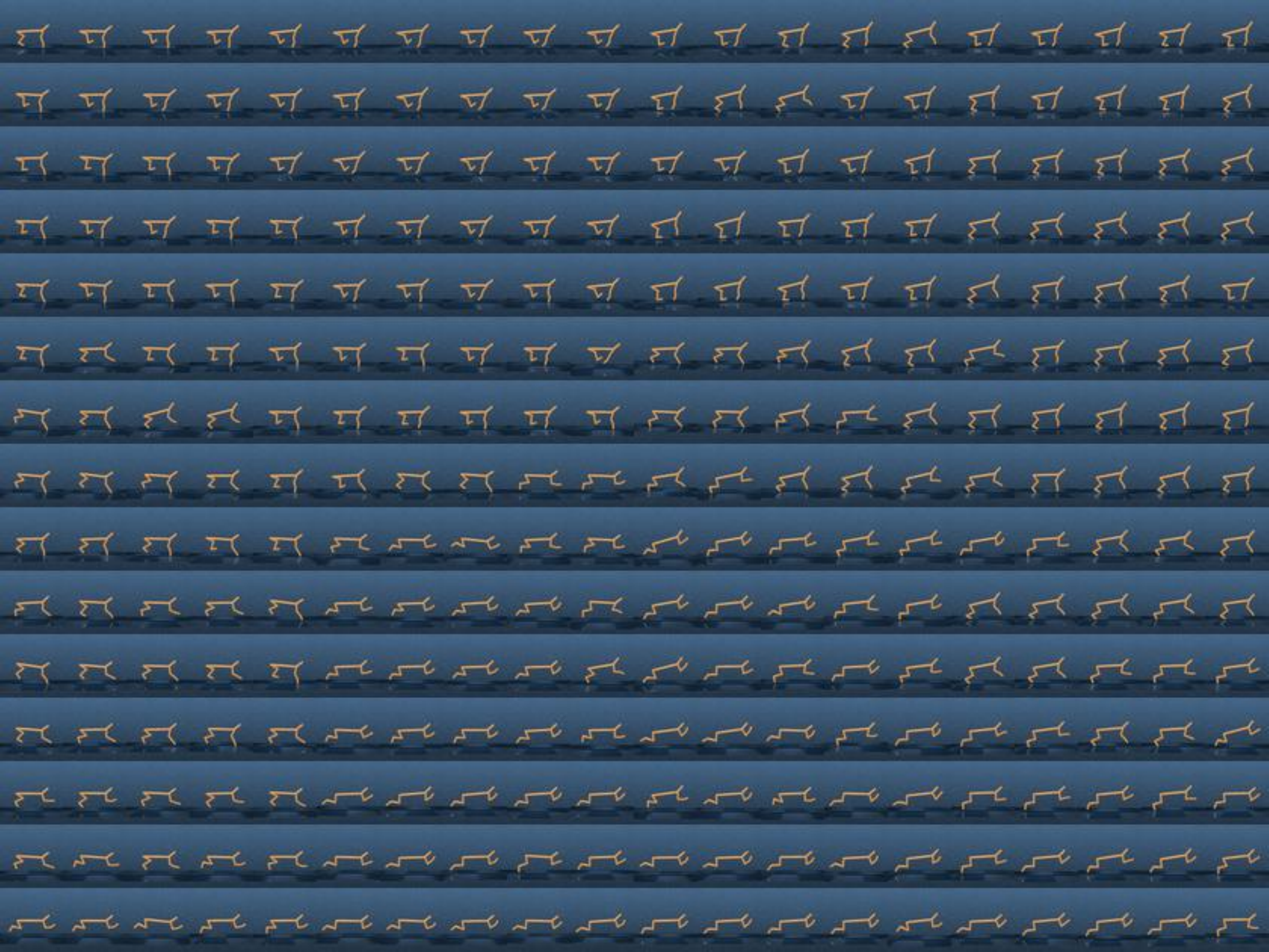}
\caption{t-SNE Visualization of representations learned with CoDy after training has completed on Cheetach Run task. The grid size is $20 \times 15$.}
\label{vis:cheetah}
\end{figure*}

\begin{figure*}
\centering
\includegraphics[width=\textwidth]{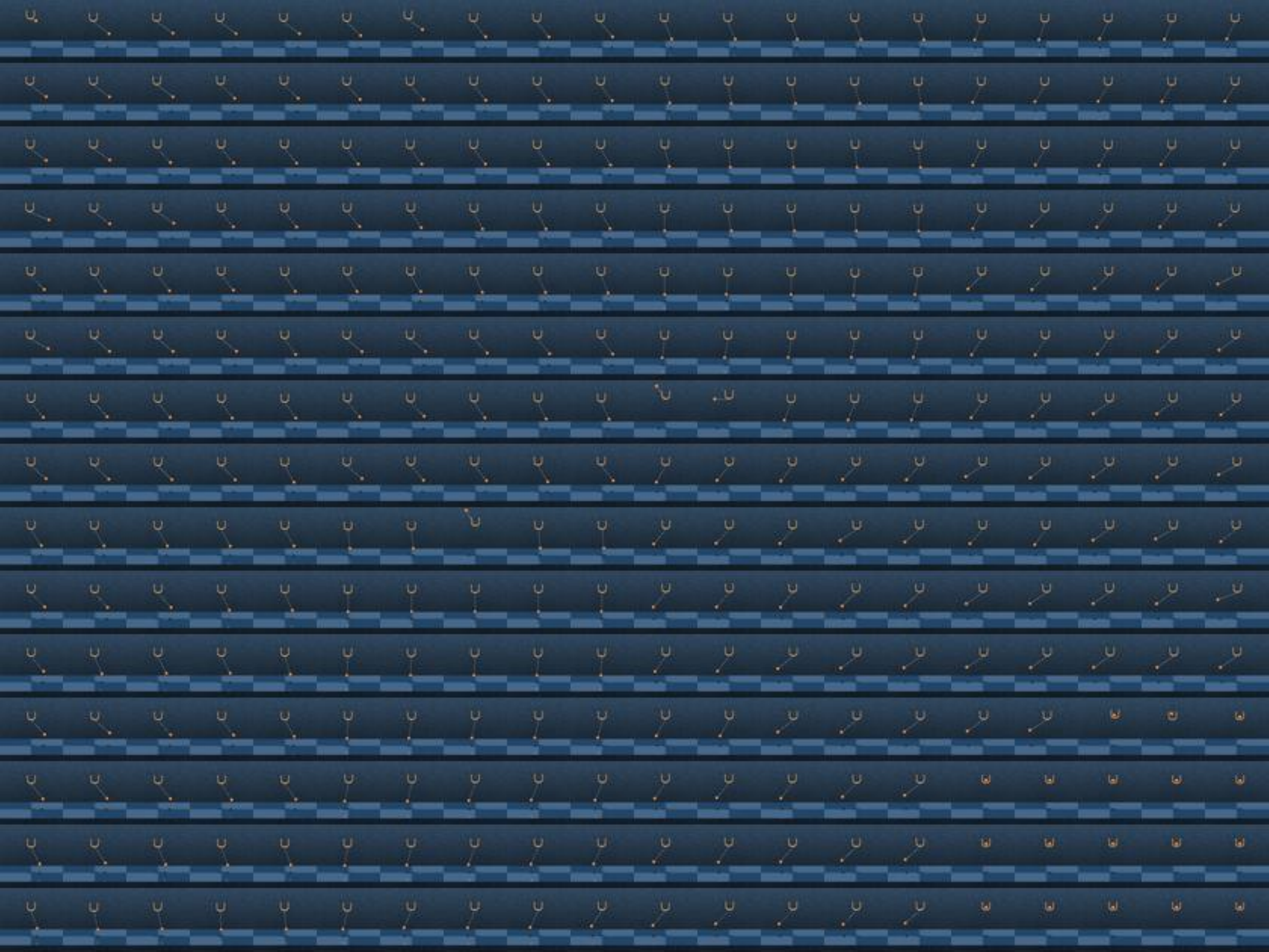}
\caption{t-SNE Visualization of representations learned with CoDy after training has completed on Ball-in-cup Catch task. The grid size is $20 \times 15$.}
\label{vis:ball-in-cup}
\end{figure*}

\begin{figure*}
\centering
\includegraphics[width=\textwidth]{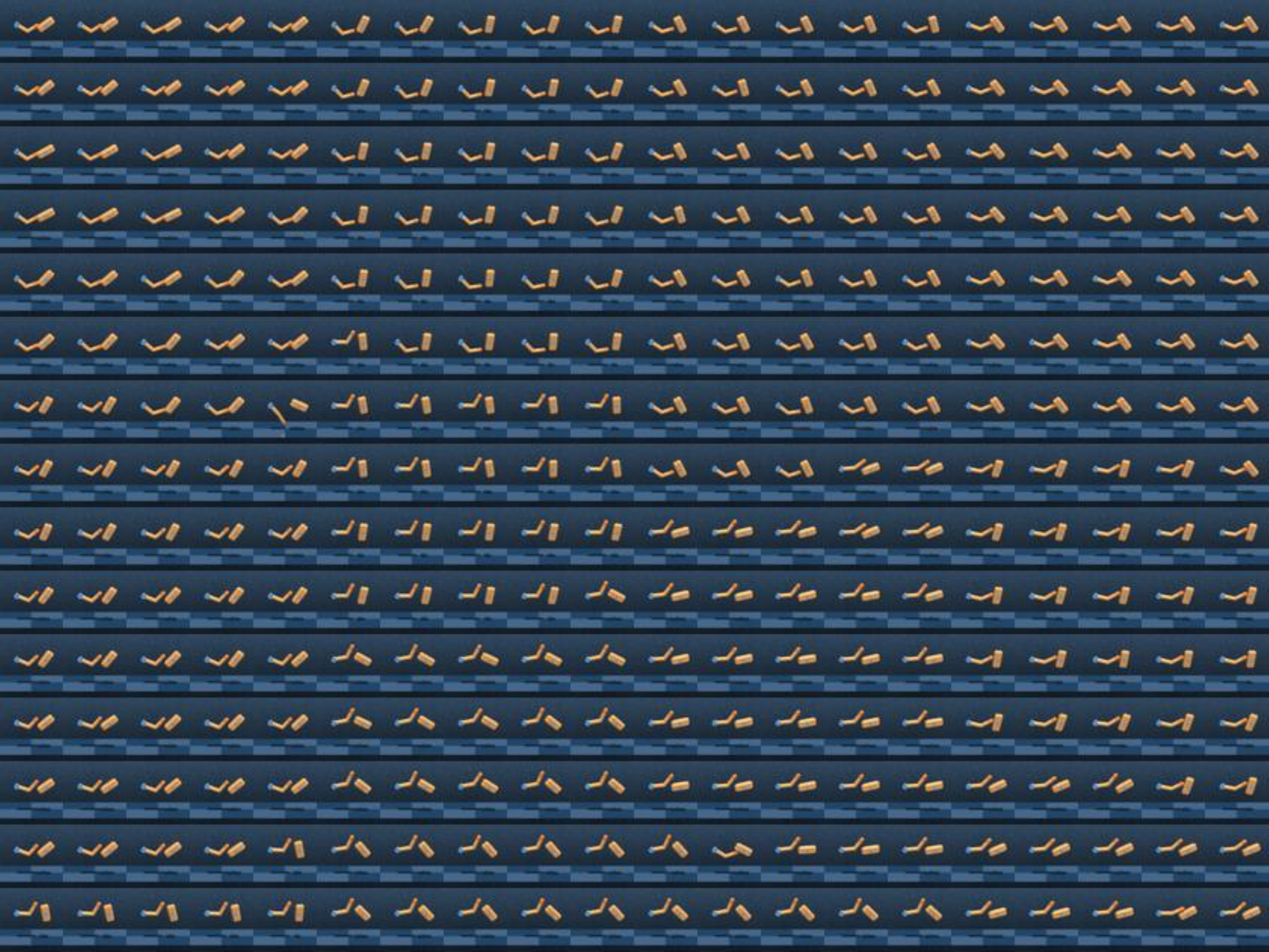}
\caption{t-SNE Visualization of representations learned with CoDy after training has completed on Finger Spin task. The grid size is $20 \times 15$.}
\label{vis:finger}
\end{figure*}

\begin{figure*}
\centering
\includegraphics[width=\textwidth]{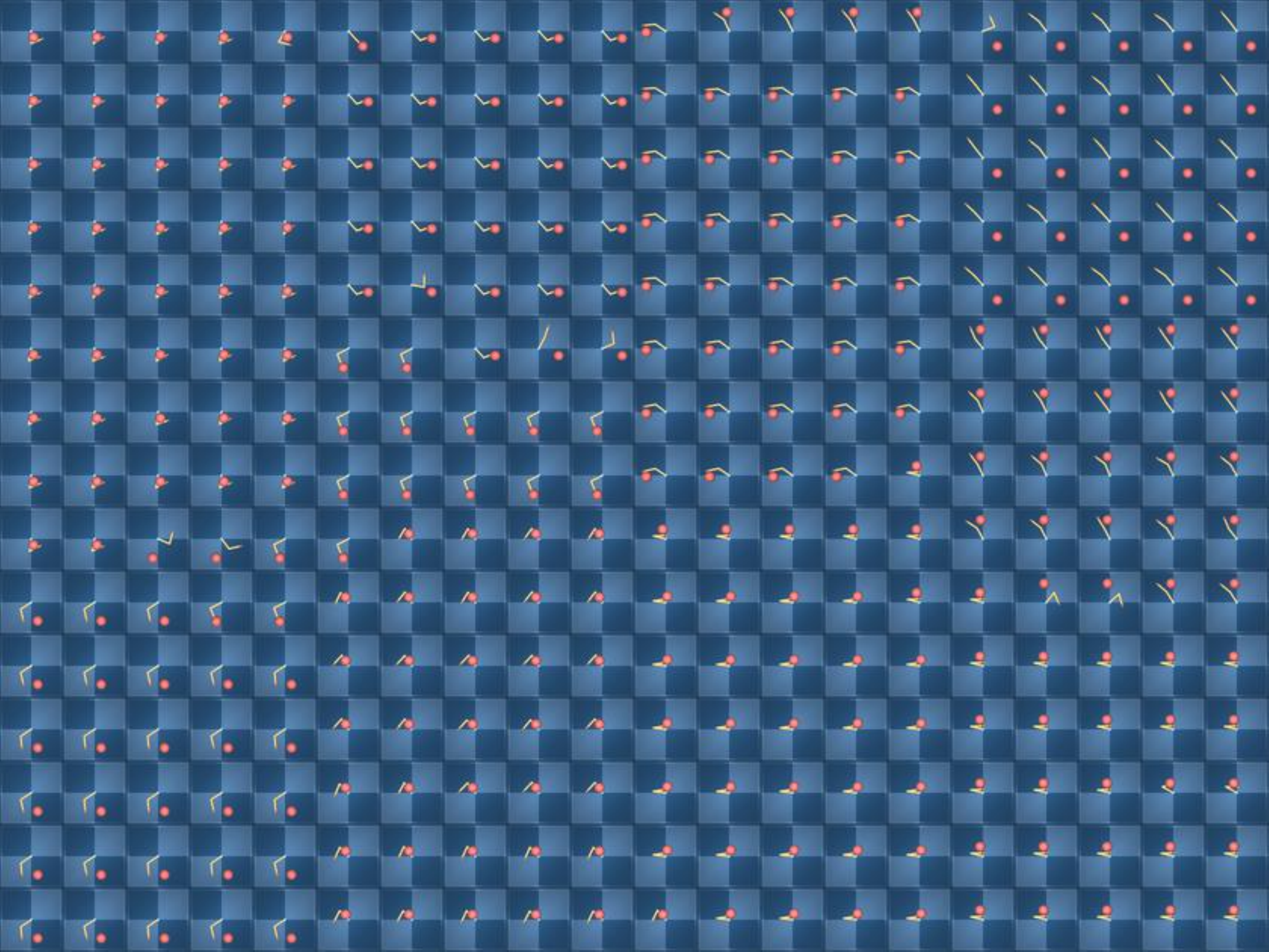}
\caption{t-SNE Visualization of representations learned with CoDy after training has completed on Reacher Easy task. The grid size is $20 \times 15$.}
\label{vis:reacher}
\end{figure*}

\begin{figure*}
\centering
\includegraphics[width=\textwidth]{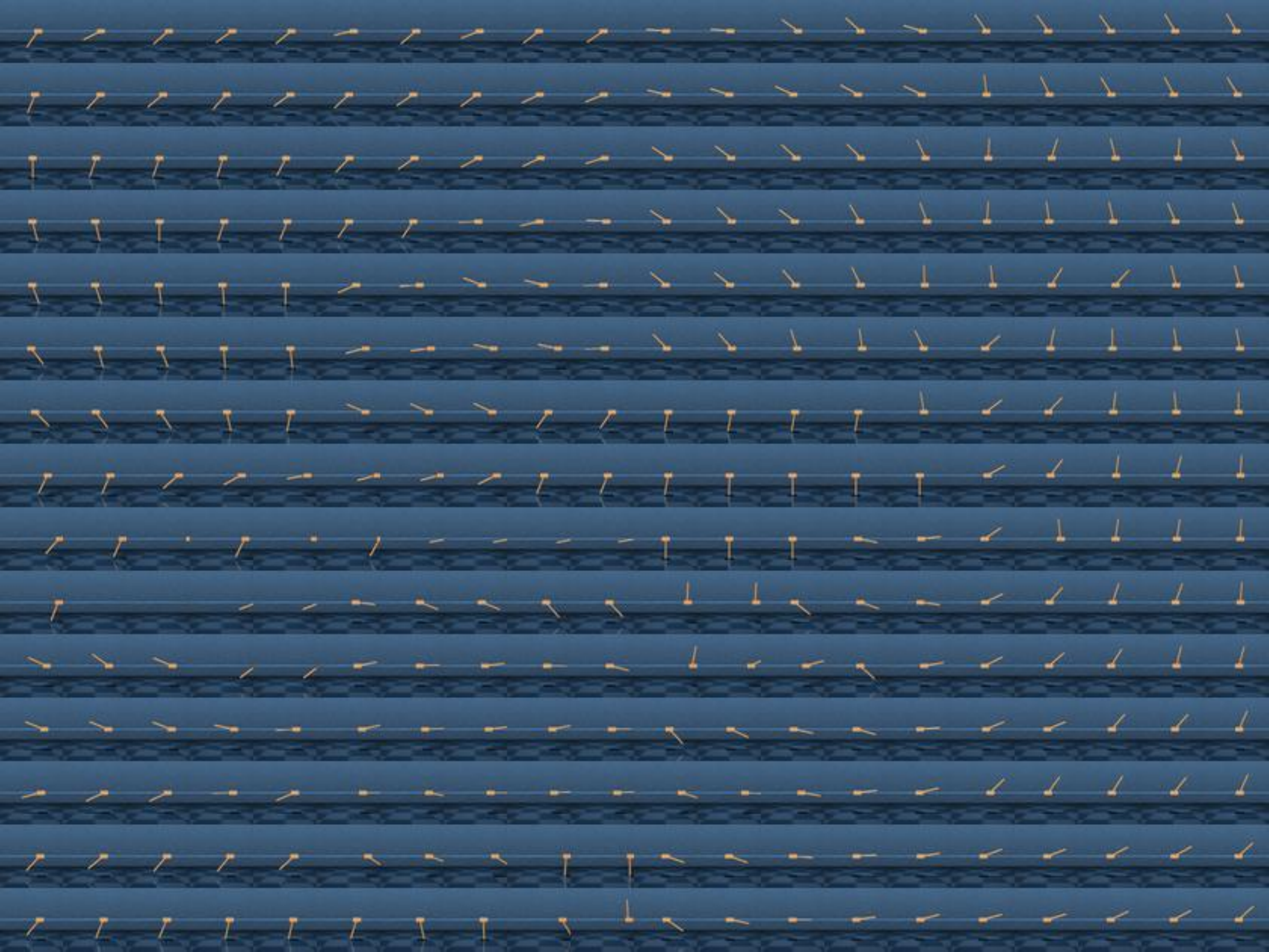}
\caption{t-SNE Visualization of representations learned with CoDy after training has completed on Cartpole Swingup task. The grid size is $20 \times 15$.}
\label{vis:cartpole}
\end{figure*}

Additional results for the experiment in Section \ref{sec:vis} that demonstrate that the latent space learned by CoDy meaningfully organizes the variations in robot configurations are shown in Figure \ref{vis:cheetah}-\ref{vis:cartpole}.

\section{Computation Time}
In order to evaluate whether our algorithm increases the time cost while improving the sample efficiency, we test computation time of our method and all baselines during policy learning. Specifically, we use a single NVIDIA RTX 2080 GPU and 16 CPU cores for each training run and record training time per 1000 environment steps after initializing the replay buffer. Table~\ref{table:computation cost} compares our method against PISAC, CURL, SAC-AE, Dreamer and Pixel SAC with respect to wallclock time. Pixel SAC has minimum time cost on all tasks among these methods, since Pixel SAC operates directly from pixels without any auxiliary tasks or learned dynamic model. Besides, we observe that the time cost of each method highly depends on the chosen hyperparameters, especially the batch size. This may explain why our method requires less training time than CURL as well as more training time than Dreamer and SAC-AE, as the batch size of CoDy is smaller than CURL's and larger than Dreamer's and SAC-AE's. Moreover, we notice that the computation time of each method is highly related to the amount of action repeat across different tasks. Hence, we assume that most of the training time is spent on gradient updates, rather than image rendering of simulated environments. When PISAC has the same amount of action repeat and batch size as our method, it has on the majority of tasks a higher computation cost than CoDy, since a single sample of PISAC contains more image frames. 

\begin{table*}[t]
\begin{center}
\sisetup{%
            table-align-uncertainty=true,
            separate-uncertainty=true,
            detect-weight=true,
            detect-inline-weight=math
        }
\resizebox{\textwidth}{!}{\begin{tabular}{c | c c c c c c}
    \Xhline{2\arrayrulewidth}
     & CoDy(Ours) & PISAC & CURL & SAC-AE & Dreamer & Pixel SAC\\
    \hline
    Cartpole Swingup &  14$\pm$ 1 & 21 $\pm$ 1 & 25 $\pm$ 1 & 10$\pm$ 0 & 18$\pm$ 1 & \bfseries 5$\pm$ \bfseries 0\\
    \hline
    Ball-in-cup Catch & 28$\pm$ 0 & 47 $\pm$ 1 & 50$\pm$ 0 & 19 $\pm$ 0  & 18$\pm$ 1 & \bfseries 9$\pm$ \bfseries 0\\
    \hline
    Finger Spin & 56$\pm$ 1 & 80 $\pm$ 1 & 98$\pm$ 0 & 37 $\pm$ 0  & \bfseries 18$\pm$ \bfseries 0 & \bfseries 18$\pm$ \bfseries 0\\
    \hline
    Walker Walk & 56$\pm$ 1 & 80 $\pm$ 1 & 98$\pm$ 1 & 38 $\pm$ 0  & 19$\pm$ 0 & \bfseries 18$\pm$ \bfseries 0\\
    \hline
    Reacher Easy & 28$\pm$ 1 & 43 $\pm$ 1 & 50$\pm$ 0 & 19 $\pm$ 0  & 19$\pm$ 0 & \bfseries 9$\pm$ \bfseries 0\\
    \hline
    Cheetah Run & 51$\pm$ 0 & 38 $\pm$ 1 & 49$\pm$ 1 & 19 $\pm$ 1  & 19$\pm$ 1 & \bfseries 9$\pm$ \bfseries 0\\
    \Xhline{2\arrayrulewidth}
\end{tabular}}
\end{center}
\caption{Computation time in seconds per 1000 environment steps(mean and standard error for 5 seeds) comparison to existing methods on six tasks from the Deepmind control suite. Bolded font indicates minimum mean among these methods.}
\label{table:computation cost}
\end{table*}

\bibliography{main}

\end{document}